\documentclass[sigconf]{acmart}
\pdfoutput=1
\renewcommand\footnotetextcopyrightpermission[1]{} % removes footnote with conference information in first column
\usepackage{blindtext, graphicx}
\usepackage{mathtools}
\usepackage{stfloats}
\usepackage{enumitem}

\setcopyright{rightsretained}

\acmDOI{10.475/123_4}

\acmISBN{123-4567-24-567/08/06}

\acmConference[WWW'18]{WWW}{April 2018}{Lyon, France} 
\acmYear{2018}
\copyrightyear{2018}

\acmPrice{15.00}

\begin{document}

\title{Facebook Reaction-Based Emotion Classifier as Cue for Sarcasm Detection}

\author{Po Chen Kuo}
%\authornote{Dr.~Trovato insisted his name be first.}
%\orcid{1234-5678-9012}
\affiliation{%
  \institution{Department of Computer Science, National Tsing Hua University}
  \streetaddress{GuangFu Rd. Sec.2, No. 101}
  \city{Hsinchu} 
  \state{Taiwan} 
  \postcode{30013}
}
\email{chuck82521@gmail.com}

\author{Fernando H. Calderon Alvarado}
\authornote{The secretary disavows any knowledge of this author's actions.}
\affiliation{%
  \institution{Social Networks and Human Centered Computing Program, Taiwan International Graduate Program, Institute of Information Science, Academia Sinica}
  \streetaddress{128, Academia Road, Sec. 2., Nangang}
  \city{Taipei} 
  \country{Taiwan} 
  \postcode{115}
}
\email{s104162862@m104.nthu.edu.tw}

\author{Yi-Shin Chen}
\authornote{Associaite Professor}
\affiliation{%
  \institution{Institute of Information Systems and Applications, National Tsing Hua University}
  \streetaddress{GuangFu Rd. Sec.2, No. 101}
  \city{Hsinchu}
  \country{Taiwan}
  \postcode{30013}
}
\email{yishin@gmail.com}

\begin{abstract}
Online social media users react to content in them based on context. Emotions or mood play a significant part of these reactions, which has filled these platforms with opinionated content. 
Different approaches and applications to make better use of this data are continuously being developed.
However, due to the nature of the data, the variety of platforms, and dynamic online user behavior, there are still many issues to be dealt with. It remains a challenge to properly obtain a reliable emotional status from a user prior to posting a comment. 
This work introduces a methodology that explores semi-supervised multilingual emotion detection based on the overlap of Facebook reactions and textual data. With the resulting emotion detection system we evaluate the possibility of using emotions and user behavior features for the task of sarcasm detection.
More than 1 million English and Chinese comments from over 62,000 public Facebook pages posts have been collected and processed, conducted experiments show acceptable performance metrics.
\end{abstract}

\maketitle
\section{Introduction}
\label{sec:intro}

Social media platforms have for long been regarded as a rich data source, especially since it is possible to understand opinions and emotions expressed in them towards a particular subject or object.
Facebook has been for some time now one of the leading online social networks~\cite{wilson2012review}.
%Its permanent development makes them add different options continuously.
One of the platform's features, Facebook Pages, provides an adequate setting for subjective comment behavior mentioned above.
Pages are in essence official accounts for an individual, media, or organization.
Posts made in these pages receive more views and comments than regular user accounts, and now, due to one of the latest additions, more reactions.
Reactions allow users not only to comment but also to express a series of emotions, in addition to the Like button.
This has made Facebook Pages posts a place laced with opinions, emotions, and sarcasm.
It is important to understand how these interact together.
It can be said that sarcasm is used to invert emotions, but conversely, can inverted emotions be an indicator of sarcasm?

% \begin{figure}[t!]
% \centering
% \includegraphics[scale=0.45]{facebook-reactions}
% \caption{Facebook reactions used as noisy labels on comments for emotion detection.}
% \label{fig:reactions}
% \end{figure}

Sarcasm has been challenging the sentiment analysis community for a while now~\cite{feldman2013techniques,maynard2014cares,farias2016irony}.
The existence and permanent proliferation of so-called internet trolls together with multiple platforms for them to operate in has not made this challenge easier.
It thus becomes useful to devise systems that can automatically detect sarcastic intentions in texts.

\begin{figure}[b!]
\centering
\includegraphics[scale=1]{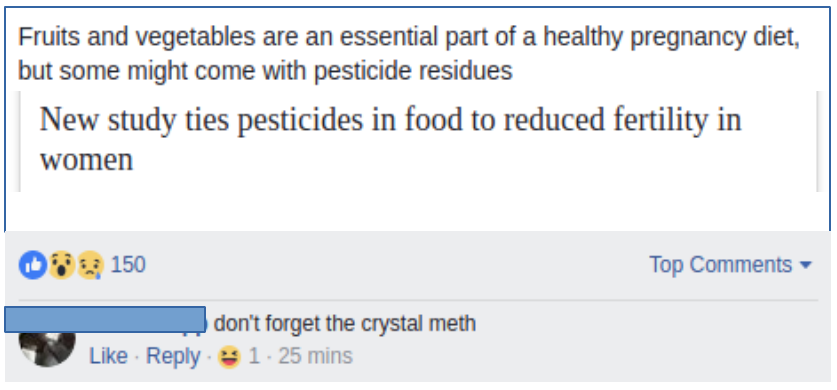}
\caption{Example of a sarcastic comment on a news post.}
\label{fig:example}
\end{figure}

When trying to detect sarcasm, context plays an important role, which implies an understanding of several factors in the setting of a comment~\cite{bamman2015contextualized,joshi2015harnessing}.
The topic, background of the user, background of the receiver, and emotions conveyed can provide some insight when determining if a comment is sarcastic or not as can be observed in Figure~\ref{fig:example}.
The main problem is that not all of these factors are available when attempting to train an algorithm to detect sarcasm.
Moreover, even if these contextual cues are available to a human reader, sarcasm may still be hard to detect.
It is thus helpful to look into other features that can provide clues or hints on the presence of sarcasm.

Some of these features may be related to the social media platforms themselves.
For example, what is the role of anonymity when incurring in sarcastic commenting? This may have an impact.
A user might feel more comfortable being sarcastic on a public page full of strangers than on a private chat with close friends.
Other users may also prefer it the other way around.
What seems to be certain, however, is that sarcasm is more prevalent as a retaliatory move than as an initial exchange.
It is more likely to find a sarcastic reply or comment than a sarcastic post, that is, as a response to try to outsmart an original post~\cite{mueller2016positive}.

This work tries to make use of the inherent characteristics of the Facebook platform with the objective of developing a system for emotion detection based only on the content extracted without the need of external knowledge.
In essence, it leverages the wisdom of the crowds in order to achieve its goal.
First, it uses the intersection of reaction clicks and comments as a fuzzy labeling technique with distant supervision where the reactions become the labels of the comments to train our emotion classifier.
It is then explored if the presence of opposing emotions with regards to a comment is an indicator of sarcasm.
Experiments have been performed for both English and Chinese comments and an extended evaluation for English is presented.
To the best of our knowledge, the self-reported reactions have not been previously used as emotion signals for labeling, nor has the possibility of detecting sarcasm from this kind of classifier been explored before.

\section{Related Work}
\label{sec:Related}

\subsection{Sentiment Analysis on Social Media}
\label{subsec:Sentiment}

Online social media platforms have increasingly attracted more interest for the sentiment and emotions expressed in their users' opinions.
This has led to the inclusion of several explicit means to reflect such sentiments in a comprehensive, user-friendly and collectible way.
Several works have focused on using these signals as noisy labels for sentiment classification~\cite{go2009twitter, davidov2010enhanced, zhao2012moodlens, hu2013unsupervised, argueta2016multilingual, wang2016sentiment}.
The work by Go et al.~\cite{go2009twitter} evaluated the performance of popular machine learning algorithms when using emoticons in tweets as labels for training via distant supervision.
In a similar way, Davidov et al.~\cite{davidov2010enhanced} leveraged Twitter features for sentiment learning and not only considered emoticons as labels, but also added hashtags.

The previous works confirmed at the time that social media could provide not only data, but annotated data that could avoid the time- and resource-intensive task of manual annotation.
This advantage was further explored by Zhao et al.~\cite{zhao2012moodlens} using data from a different platform (Weibo) and language (Chinese).
Their system, MoodLens, maps 95 emoticons into 4 sentiment classes and became one of the pioneering tools for sentiment analysis from short texts in Chinese.
Another study using the Weibo platform performed sentiment correlation to determine which of two emotions---anger and joy--is more influential in a social network~\cite{fan2014anger}.
Lipsman et al.~\cite{lipsman2012power} focused uniquely on the number of Likes in a post to determine what kind of repercussions this click behavior had from the perspective of brands and fans.

Following a similar trend, the work by Hu et al.~\cite{hu2013unsupervised} studied the use of emotion signals not only as labels for training, but also as an active part in unsupervised learning models for sentiment analysis.
Hashtags on its own have also been used for similar tasks.
Argueta et al.~\cite{argueta2016multilingual} used hashtags for distant supervision on unsupervised methods to collect writing patterns that can be correlated to emotions.
It has been found that using emoticons or hashtags as labels can lead to some errors.
This served as motivation for Wang et al.~\cite{wang2016sentiment}, who proposed a method for ``de-noising'' the obtained labels.

Despite the availability of multiple online social networks, most of the related work has been focused on Twitter.
Ortigosa et al.~\cite{ortigosa2014sentiment} were among the first to perform sentiment analysis on Facebook.
Their application---SentBuk---tries to help e-learning systems by providing sentiment information of users through their posts.
The achieved performance shows that Facebook data can also be used for sentiment related tasks.

Recent years have witnessed the development of algorithms that deliver very high performance on sentiment related tasks.
VADER, the rule-based model developed by Hutto and Gilbert~\cite{hutto2014vader}, tries to make the most out of sentiment lexicons combined with machine learning algorithms.
Deep convolutional neural networks have also had a significant participation in sentiment classification tasks.
The work by~\cite{poria2015deep} presents a model for multi-modal classification of short sentences based on features extracted from text.
As highlighted by Liu~\cite{liu2015sentiment} and Cambria~\cite{cambria2016affective}, however, there are several factors weighing in on sentiment related topics, many of which have not been thoroughly explored.
For instance, Volkova et al.~\cite{volkova2013exploring} tried to explore the impact of demographic language variations when attempting multilingual sentiment analysis, and made clear how this can be an issue.
The context on which opinions are expressed is also of high importance, as studied by Muhammad et al.~\cite{muhammad2016contextual}.
Their work explains that depending on the social media genre being studied, significant variations in the modeling are required.
The impact of innate human responses such as sarcasm is also one of the factors pending an in-depth exploration.

\subsection{Sarcasm Detection}
\label{subsec:Sarcasm}

Sarcastic expressions are a natural product of humor improvisation and have found a natural proliferation space in online social media.
With them, they bear a lot of trouble for mining tasks due to the uncertainty and ambiguity they bring to expressions.
If it is hard for humans to define and identify sarcasm, it is even harder to teach a computer how to do it.
Nevertheless, the research community has done efforts to achieve this.

Maynard and Greenwood~\cite{maynard2014cares} highlight the importance of understanding the impact of sarcasm in sentiment analysis.
Reyes et al.~\cite{reyes2012humor} first attempted to identify humor and irony, as this could provide some insights to sarcastic expressions.
Based on textual features and leveraging on the hashtags \emph{\#humor} and \emph{\#irony}, they developed a system to identify ``figurative language''.
Bamman and Smith~\cite{bamman2015contextualized} believe that sarcasm is a highly contextual phenomenon and that extra-linguistic information is required for its detection.
They consider lexical cues and their corresponding sentiment as contextual features in their study.
Rajadesingan et al.~\cite{rajadesingan2015sarcasm} go beyond these affirmations and claim behavioral traits are also intrinsic to users expressing sarcasm.
They developed a model for sarcasm detection based on the analysis of past tweets paired with behavioral and psychological studies. Riloff et al.~\cite{riloff2013sarcasm} attempts to identify situations in which a situation and its subsequent reaction have opposing sentiment polarities. They use this as a clue to identify sarcastic expressions using bootstrap learning methods. Gonzalez-Ibanez et al.~\cite{gonzalez2011identifying} also experimented with the sentiment polarity in twitter messages and the presence of sarcasm transforming this polarity. Their work uses lexical and pragmatic features to train a machine learning system to identify these utterances. Lexical feature where also used by Bharti et al.~\cite{bharti2015parsing} in developing their  parsing-based lexicon generation algorithm to detect sarcasm on twitter.

Sarcasm detection has also been attempted in other languages.
The work by Lunando and Purwarianti~\cite{lunando2013indonesian} first performs sentiment classification on short texts from Indonesian social media.
It then considers two other features: negativity information and the number of interjection words to perform sarcasm detection through machine learning algorithms.
Liebrecht et al.~\cite{liebrecht2013perfect} built a Twitter-based corpus by collecting tweets containing the hashtag \emph{\#sarcasm} and trained a machine learning classifier.
The data used was in Dutch, but still showed that sarcasm is often signaled by intensifiers and exclamation marks.
Tweets in English and Czech were studied by Ptacek et al.~\cite{ptacek2014sarcasm} to develop a language-independent approach borrowing features across languages. The work by Liu~\cite{liu2014sarcasm} explores sarcasm detection on Chinese text primarily focusing on the issue of imbalanced data and proper feature selection which is evaluated through multiple classifiers. The dataset used contain simplified Chinese characters, to the best of our knowledge our work is the first to address this problem for traditional Chinese characters.

\section{Methodology}
\label{sec:method}

\begin{figure*}[htp!]
\centering
\includegraphics[scale=0.55]{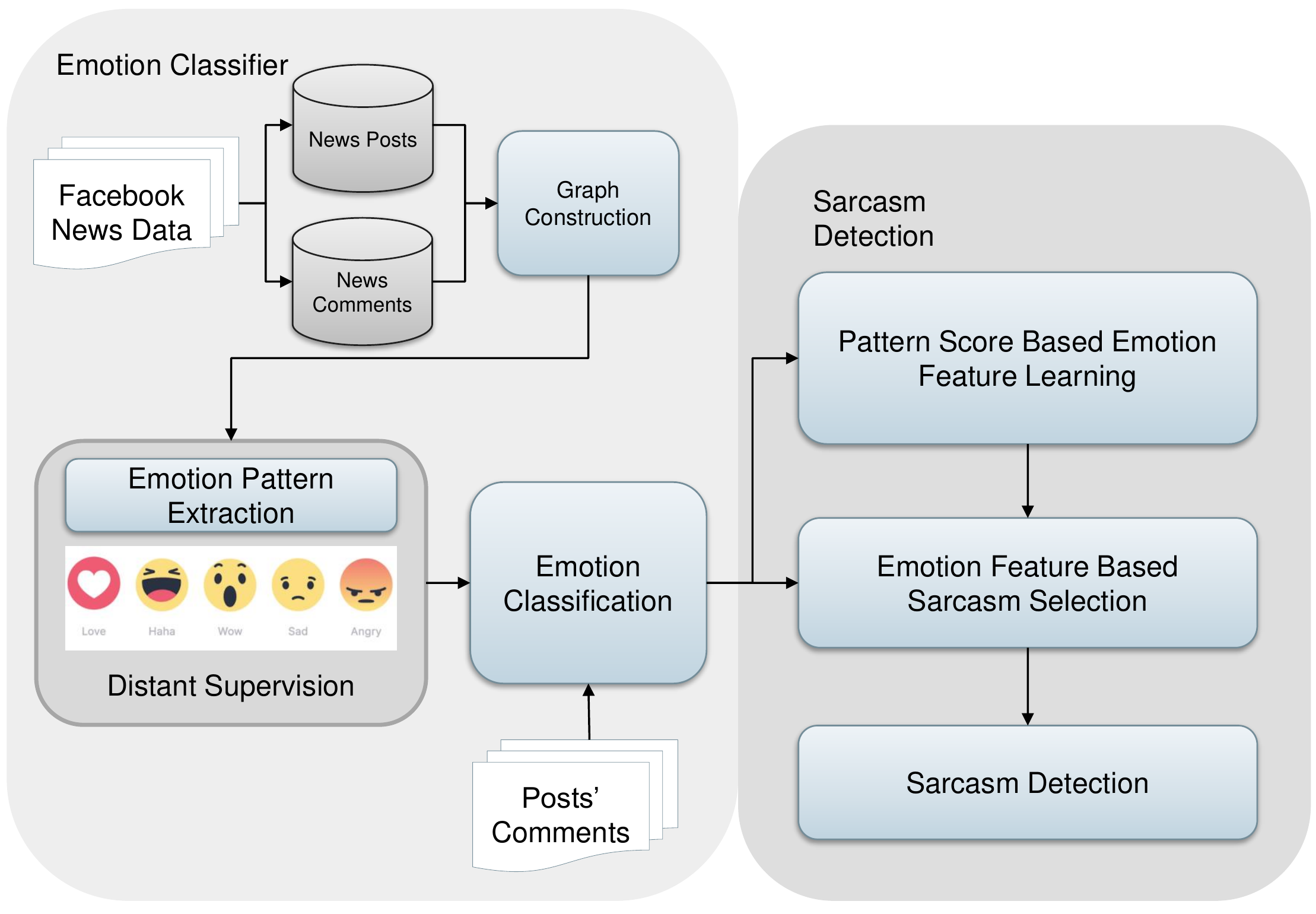}
\caption{Methodology flowchart.}
\label{fig:flowchart}
\end{figure*}

\subsection{Overview}
\label{subsec:Overview}
Taking from some of the approaches mentioned in the related work, emotion classification is first performed on short texts.
The training data and labels are obtained from the intersection between reaction clicks and comments from users corresponding to those reactions.
The classifier returns the two most likely labels corresponding to a text.
These two labels will then undergo feature evaluations that will help determine if a comment is sarcastic or not.
A flowchart for the method is presented in Figure~\ref{fig:flowchart}.

\subsection{Data Collection}
\label{subsec:datacol}

One of the key features of this work is the exploitation of embedded characteristics of the Facebook platform, the first being their ``Pages'' feature.
Facebook Pages are official accounts of a varied types of sources, popular personalities, organizations, and media outlets.
Our methodology takes particular advantage of the official pages of news media.
The implemented emotion classification algorithm requires objective and subjective data in its development.
By using pages from news media, objective texts can be obtained from news posts, while subjective texts can be obtained from user comments.
It is intuitive that comments on these articles are usually highly opinionated, sometimes biased, and predominantly subjective.

The second key element to be used is Facebook Reactions.
Since the beginning of 2016, the traditional ``Like'' button was replaced by a more variety-aware option called Reactions.
Reactions are emoji-based expressions that allow a user to express their sentiment towards a post.\footnote{Even though some of these reaction emojis are not emotions in their strict sense (as listed in Plutchik's wheel of emotions~\cite{plutchik1980emotion}), they may nevertheless provide some insight into a user's sentiments.}.
It was identified that many of the users who react to posts also have a tendency to comment on them.
The proposed data collection approach is to find the intersection between reaction clicks and comments that will enable a matching between a user comment and its corresponding reaction.
This not only allows a filtering process to build a collection, but also guarantees that there will be a self-reported emotion assigned to every comment, which basically results in an automatic annotation.

The previously mentioned characteristics enable the collection of objective news data and subjective comments data, the latter which is paired with emotion labels.
This meets the requirements for the implementation of the pattern-based emotion classifier to be used in this work.

\subsection{Reaction-Based Emotion Classification}
\label{subsec:emotion}

The scope of this work requires a data driven approach with multilingual capability. This goes in line with the emotion classifier proposed by Argueta et al.~\cite{argueta2016multilingual} which is taken as a reference for the classifier introduced in this work.
In general, the system builds a graph with subjective terms from short text.
It then extracts patterns of expression from this graph and assigns weights to them across a multiple range of emotions.
Being pattern based, it allows for multi-lingual analysis.
The implementation used in this work varies from the original in the following aspects:
\begin{itemize}
  \item Facebook data is used, rather than Twitter data.
  \item Chinese language is integrated.
  \item Emotions from Facebook Reactions are used as labels instead of Plutchnik's emotion set.
  \item Reaction-Comment pairs are considered as annotations rather than Hashtag-Tweet pairs.
\end{itemize}

\subsubsection{Graph Construction}
\label{subsubsec:graphc}

\begin{figure}[H]
	\includegraphics[scale=0.6, trim=20 10 10 20]{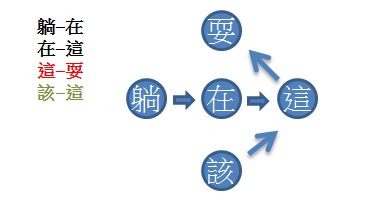}
	\includegraphics[scale=0.5, trim=20 20 10 20]{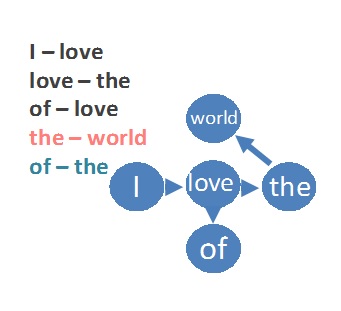}
	\caption{Examples of subjective graphs from Chinese and English comments.}
	\label{fig:subj_graphs}
\end{figure}

The data obtained, as defined in the previous section, is converted into graph form for further manipulation.
The nodes in the graph correspond to words, and the edges denote the co-occurrence between the connected words. The order in which words appear is also considered in the co-occurrence and hence reflected by the direction of the edges.
The graph obtained from the news posts intuitively contains more factual, objective expressions, while the graph from the comments is more subjective and opinionated.
Figure~\ref{fig:subj_graphs} shows examples of graphs constructed from comments (i.e., subjective) in Chinese and English.

\begin{figure}[H]
	\includegraphics[scale=0.40, trim=140 0 100 10]{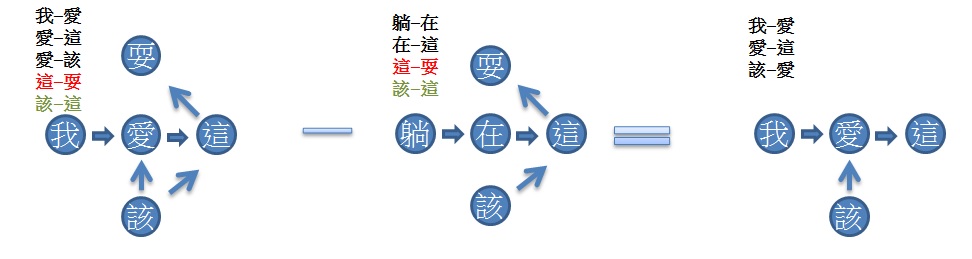}
	\includegraphics[scale=0.40, trim=140 30 100 0]{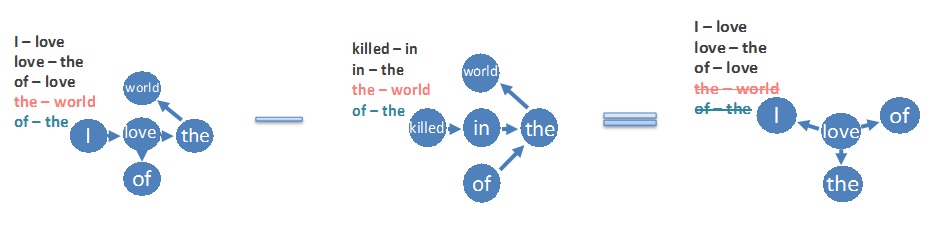}
	\caption{Examples of graph reduction for Chinese and English graphs.}
	\label{fig:graph_reduction}
\end{figure}

The next goal is to obtain a set of expressions that are highly subjective so that they can be linked to particular emotions.
With this purpose, a reduction is performed on the comments graph by removing terms that are highly dominant in the news graph.
This procedure reduces the objective components present in the comments graph, resulting in a highly subjective graph.
An example of graph reduction is shown in Figure~\ref{fig:graph_reduction} for both Chinese and English graphs.

The graph-construction process for Chinese text poses an additional difficulty, given that it requires more steps in its pre-processing, particularly in word segmentation.
In Chinese language, a word can be composed generally by one, two or three characters.
Characters on their own may have a meaning when appearing alone, and they can mean something totally different when paired with another character.
It is therefore of high importance to perform appropriate character segmentation and sense disambiguation before proceeding to build the graph.
In order to segment Chinese characters, the following hierarchical combination is performed:
\begin{enumerate}
	\item Combine the characters into two-, three-, and four-character words and calculate their frequency in the dataset.
	\item Perform an initial reduction on three-character words by subtracting the frequency of four-character words that contain them.
	\item Perform an additional reduction on two-character words by subtracting the frequency of three- and four-character words that contain them.
\end{enumerate}
After the previous procedure is complete, frequent words are filtered into two-, three-, and four-character words using an arbitrary frequency threshold.

\subsubsection{Emotion Patterns}
\label{subsubsec:emopatterns}

Repetitive instances of sequences in the graph with $length=2$ or $length=3$ sharing one or two words will become patterns.
Since the graph is filled with subjective expressions, the intuition is that the obtained patterns are expressions that denote a high level of emotion.
It is also important to determine which emotion a pattern is more likely to be expressing.

 \theoremstyle{definition}
    \begin{definition}
    An element \(e\) is a word or a sequence of symbols (,.?!, etc).
    \end{definition}
    \begin{definition}

    A pattern $P_i$ is a sequence of two or three elements.
    %\begin{center}
    %    {T}  represents any message in social media.
    %\(ws\) is a word sequence in text.
    %\[{ws}=\{{w_1,w_2,w_3}\}\in {T}, \forall{w}_i \in {ws} \text{,is a word}.\]
    %\[\mathcal{EW} \text{ is a subset of emotion words defined in LIWC.}\]
    %\[{ws_x}\text{ is a word sequence without emotion words.}\]
    %\[{ws_x} \cap \{{EW}\} = \emptyset\]    
    %\[\{{EMO}\}\text{ is a set of emotions.}    \]
    %\[\{{EMO}\}=\{{emo_1,emo_2,...,emo_n}\}=\{{Haha, Angry, Sad, Love, Wow}\}\]
    %\[{getEmo()} \text{ is an emotion classification function.}\]
    %\[getEmo({ws_x})=emo_i\]
    %\end{center}
    \begin{equation}    
    {P_i} = [e_1, e_2, e_3] 
    \forall {P_i} \in \mathcal{P}
    \end{equation}
    \end{definition}

The obtained emotion patterns are then paired to our set of labels through distant supervision.
Given that we have our set of patterns and a set of comments with their corresponding emotion labels, the idea is to find how many instances of the patterns are in the corpus.
By doing probabilistic analysis, it can be determined in which particular emotion label a certain pattern was more predominant.

    \begin{definition}
    An Emotion Degree \(ED(emo,p)\) is a score representing how a pattern is related to a specific emotion.
    \begin{equation}
    ED(emo,p) \longmapsto ed, ed \in \mathbb{R^+}
    \end{equation}
    \end{definition}
    
    \begin{definition}
    An emotion \(emo\) is defined in a set of 5 emotions.
  All the patterns have 5 Emotion Scores(ES) each with a corresponding emotion.
    \[
    emo \in Emotion \{Haha, Angry, Sad, Love, Wow\}
    \]
    \end{definition} 
    As a result, every emotion will contain the same patterns, but ranked in a different order and weighed by Emotion Degree \(ED\) that depends on their frequency, uniqueness, and diversity.
    
    \begin{definition}{Pattern Frequency (PF)}
    
    Pattern Frequency \(PF(emo,p)\) represent the frequency of an emotion pattern \(p\) in an collection of social data related to emotion \(emo\).
    \begin{equation}
    PF(emo,p) = log (f(p,emo) + 1)
    \end{equation}
    \end{definition}
    
    \begin{definition}{Inverse Emotion Frequency (IEF)}
    
    The Inverse Emotion Frequency \(IEF(emo,p)\) is a measurement of how rare or unique a pattern \(p\) is across all emotion classes.
    \begin{equation}
    IEF(emo,p) = \frac{|Emotion|}{|\{emo \in Emotion : f(p,emo) > 0\}|}
    \end{equation}
    \end{definition}
    
    \begin{definition}{Diversity (DIV)}
    
    Diversity \(DIV(p)\) consider the number of unique psychology words
    (denoted as \(uew\) that fit the pattern \(p\) across all emotion classes \(emo\).
    \begin{equation}
    DIV(p) = log (uew(p,Emotion))
    \end{equation}
    \end{definition}
    %\begin{itemize}
    %\item Pattern Frequency (PF) : the frequency of the patterns with normalization in each emotion.
    
    %\item Inverse Emotion Frequency (IEF) : similar to the inverse document frequency, it's a measurement to the importance of the pattern across all emotion classes.
    
    %\item Diversity (DIV) : consider the number of unique psychology words with the pattern across all emotion classes. 
    %\end{itemize}
    
    Finally, the emotion degree that shows how important a pattern is in an emotion class is obtained by the equation below.
    \begin{equation}
    %\[
    ED(emo,p) = PF(emo,p) \times IEF(emo,p) \times DIV(p)
    %\]
    \end{equation}

{\small
\renewcommand{\arraystretch}{1.2}
\renewcommand\tabcolsep{8pt}
\begin{table}[htp!]
	\centering
	\caption{Examples of highly ranked patterns for some emotion labels in English.}
	\label{tab:patterns}
	\begin{tabular}{l l l l}
		\hline
		%\multicolumn{5}{c}{Examples of Patterns by Emotion Reaction}                 \\ \hline
		Angry          & Haha         & Wow         & Sad           \\ \hline
		* all haters   & * . lol      & * . awesome & * so sad      \\
		trump is *     & happy bday * & a * what    & my heart *    \\
		what a *       & * ! yeah     & * user omg  & * god bless   \\
		* this country & ever !! *    & !!!! * !    & prayers for * \\
		people are *   & looks so *   & * !!! how   & . rip *       \\ \hline
	\end{tabular}
\end{table}
}

Table~\ref{tab:patterns} contains examples of extracted patterns that are ranked high and thus very representative of the corresponding labels.
It is worth noting that the corpus is crawled from Facebook pages of news media, so at the time of the crawl, these were rich in political content, international conflicts, etc.
This can be particularly evident for Anger and Sadness, where the topic related to the corresponding comments that generated these patterns can be deduced.
Other categories also have particular characteristics.
For example, in the Wow emotion, there is a presence of question words such as ``what'' and ``how'', which can be indicators of surprise.

The presence of the wildcard(*) in the patterns is also worth noticing.
The wildcard(*) takes the place of a word that can elicit a high degree of sentiment.
These words are replaced by this token so that any word that is used in the same way can be matched by these patterns.
For instance, the pattern ``people are *'' could match ``people are dumb'' or ``people are stupid'', both denoting an angry expression; the usage of the wildcard(*) thus allows matching both examples to the same pattern.
A pattern's ability to capture many different instances is what was referred to previously as diversity.

\subsubsection{Emotion Classification}
\label{subsubsec:emoclass}
The classification process then takes a new unlabeled comment, and through a matrix multiplication procedure, it evaluates it with the patterns and ranks within the labels.
The process first determines which patterns are present in the post.
It then proceeds to calculate the score of how likely a text is to belong to a class, depending on the score and ranking of the patterns it contains.
As a result, the system returns a scored and ranked list of the emotion labels based on the likelihood of the new comment belonging to them.
For practical purposes, the top two results are considered as the labels for the evaluated text.
These two labels are then evaluated to see if they can provide insights to the presence of sarcasm.

\subsection{Sarcasm Detection}
\label{subsec:sarcasm}

Sarcasm is a highly context-dependent reaction---it is usually not planned for, but initially depends on previous information.
The post-then-comment scenario from which the data is crawled provides a kind of interaction that may favor this behavior.
For example, a user first reads a news post, and depending on his/her opinions towards the topic, s/he might decide to first react to it in a peculiar way and then provide a sarcastic reply to the post.

Another characteristic of sarcasm--- and one that has troubled the sentiment analysis community---is the reverting of a emotion from the perceptive point of view.
This is the typical use of positive statements when actually having a negative point of view.
If the receiver is not aware of the state of humor or behavioral traits of the sender, then the message may be perceived as positive, while the intention may have been negative.
This poses a significant challenge to automatic emotion classification systems, since they cannot be aware of these particular behavioral traits.
Our methodology tries to make use of the flip of emotion as a feature for sarcasm detection, as explained in the following section.

\subsubsection{Sarcasm Candidate Filtering}
\label{subsubsec:sarcasm_filtering}

Based on the aforementioned user behavior, this stage initially determines if a comment is at all eligible for containing sarcasm.
After performing emotion classification for a large set of comments, a particular case arose in which many of the results consisted of opposing sentiment labels, specifically Anger and Haha.
This is perhaps due to the nature of  the data and the presence of internet trolls in these kind of sites, which can lead to a user reacting with laughter to a piece of news that would otherwise generate anger in the majority of the population.
Nevertheless, this also relates to sarcastic behavior.

With this behavior, every short document with opposite emotions is considered a candidate for sarcasm detection. To find the combination of emotions which can indicate sarcasm, a machine learning method is used to evaluate this possibility. Our method tries to make use of combined emotions as a feature for sarcasm detection, as explained in the following process.

\begin{itemize}[leftmargin=*]
    \item Convolutional Neural Network
\end{itemize}

\begin{figure}[H]
\includegraphics[scale=0.32]{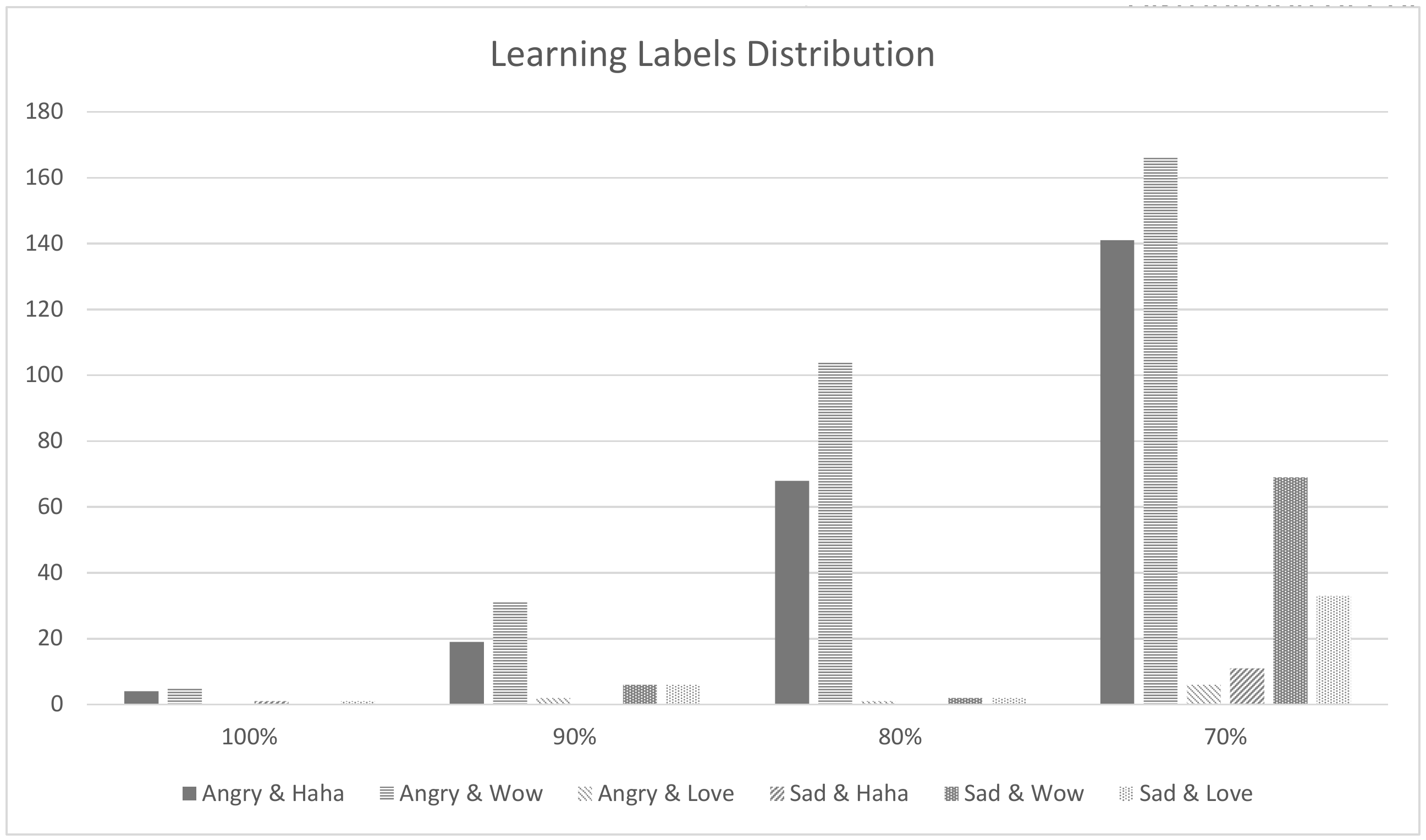}
\caption{Example of emotion combination Learning Result for English comments.}
\label{fig:learning}
\end{figure}

\begin{itemize}
    \renewcommand\labelitemi{---}
        \item Input Matrix
        
        Since every emotion will contain the same patterns weighted by a score and ranked in different order accordingly, we consider the Pattern Scores  \(PS\) of the comment in each emotion as the input matrix of the Convolutional Neural Network.
        
        Every comment evaluated will generate an input matrix as follows:
        \begin{center}
        \[
        \begin{pmatrix} 
        & Pattern Score_1 & ... & Pattern Score_n\\ 
        Emotion 1 & 300 & ... & 2000\\
        Emotion 2 \\
        Emotion 3 \\
        Emotion 4 \\
        Emotion 5 \\
        \end{pmatrix}
        \]

        \end{center}
            
        \par The Pattern Score is the ranking of the pattern in each emotion multiplied by the frequency of the pattern in the comment .Here we consider n pattern in every emotion where n is experimentally defined.
            
        \item Training Prediction
        
        When training the model, we use the training prediction as our observation source. We then calculate in how many of the iterations is a sarcastic comment correctly identified, in parallel this lets us know those that can be correctly learned by our model. We call this value the correct training rate.
        Since sarcasm has the characteristic of flip of emotion, we then consider the combinations of opposite emotions that allow a correct prediction.
        By calculating the emotion combination results for the range 100\% to 70\% correct training rate, we can extract the specific combinations which represent more precise indicators for sarcastic comments. For example, by observing the correct learning rate at 70\% we can identify which 2 combinations of emotions where more useful for the classification as illustrated by Figure~\ref{fig:learning}:
\end{itemize}
    
            %\begin{figure}[H]
            %    \includegraphics[scale=0.45]{images/Chi_cnn.pdf}
            %    \hline
            %    \caption{Chinese emotion combination Learning Result.}
            %\end{figure}

After performing this evaluation it is observed that the emotion pairs Angry \& Haha and Angry \& Wow are the most useful as learning features for sarcasm detection. It is therefore determined that every short document with these two emotion labels resulting from the classifier is thus considered a candidate for sarcasm detection.

\subsubsection{Sarcasm Labeling}
\label{subsubsec:sarcasm_label}

It was observed that just the presence of the two specific emotion labels wasn't directly an indicator of sarcasm. There is a dependency on the distance between these two initial labels. If the top label is very dominant compared to the subsequent ones there is less chance for it being a sarcastic instance. On the other hand if the two top labels have similar scores and opposing sentiment there is higher chances for it to be a sarcastic comment.
Once a sarcasm candidate is received, two measurements between its two emotion labels are obtained. These values are the determining factor in deciding if a comment is sarcastic or not, and are specified by Definition~\ref{def:dist_rat} and Definition~\ref{def:score_rat}.

    \begin{definition}{Distance Ratio Measurement}
    \label{def:dist_rat}
    
    To make sure there is not only one specific emotion, we measure the difference of emotion score of emotion 1 and 2 with emotion 2 and 3.
    We divide the difference of emotion score of emotion 1 and 2 by emotion 2 and 3.
    The value of our measurement need to greater or equal to $x_1$, and less than or equal to $x_2$ where $x_1$ and $x_2$ are experimentally defined.

    \begin{equation}
    x_2 \geq\frac{Score(emotion_3) - Score(emotion_2)}{Score(emotion_2) - Score(emotion_1)} \geq x_1 
    \end{equation}
    \end{definition}
    
    \begin{definition}{Score Ratio Measurement}
    \label{def:score_rat}
    
    To make sure there is not only one specific emotion, we measure the emotion score ratio of emotion 1 and 2 with emotion 2 and 3.
    The value of emotion score ratio of emotion 1 and 2 need to greater or equal to $y_1$, and the value of emotion score ratio of emotion 2 and 3 need to greater or equal to $y_2$ where $y_1$ and $y_2$ are experimentally defined.

    \begin{equation}
    \frac{Score(emotion_3)}{Score(emotion_2)}\geq y_1 , \frac{Score(emotion_2)}{Score(emotion_1)} \geq y_2
    \end{equation}
    \end{definition}

The first guarantees that there is not only one emotion by evaluating distance between scores. The second one if for normalization, since the ratio from one emotion to the next can be a better indicator than just the distance. It is worth mentioning that the thresholds vary across languages perhaps due to cultural differences and language expression. It was found for example that Chinese posts tend to contain one dominant emotion, with an outstanding score. Therefore the range must adjust to this characteristic.
%\begin{definition}{Sarcasm Labeling}\\
%\label{def:sarc_labeling}
%A comment is labeled as sarcastic iff %$\Delta_{Emotions}$ $\leq$ $\lambda$,\\
%where $\Delta_{Emotions}$ = $Score_{Emo1}$ - %$Score_{Emo2}$,\\
%and
%$\lambda$ is an predefined threshold.
%\end{definition}

\section{Experiments and Results}
\label{sec:exp}

\subsection{Data}
\label{subsec:data}

The collected posts come from a variety of public Facebook Pages belonging to news media outlets in both Chinese and English---both datasets were evaluated separately. A total of 62,248 posts were crawled, together with the comments and reactions contained in them. Around 46,253 posts with approximately 3 million comments  correspond to Chinese data, crawled on a period between June 1-July 31 2016. The remaining 15995 posts with approximately 7 million comments are in English and were obtained on a period between October 1-November 30  2016.
After the comments were collected, they were matched with a corresponding reaction chosen by the user.
Table~\ref{tab:counts} presents the total counts of comments overlapped to a particular reaction for both English and Chinese datasets.
These sets of comments with their self-reported annotations are used to train the system.\footnote{The collected sets and corresponding labels can be made available upon request.}

\renewcommand{\arraystretch}{1.2}
\renewcommand\tabcolsep{5.3pt}
\begin{table}[h]
\centering
\caption{Counts of collected comments per corresponding emotion for both English and Chinese.}
\label{tab:counts}
\begin{tabular}{llllllll}
\cline{1-4}
\multicolumn{1}{l}{\begin{tabular}[c]{@{}l@{}}Overlapped\\   Emotion\end{tabular}} & \multicolumn{1}{l}{\begin{tabular}[c]{@{}r@{}}Chinese\\   Comments\end{tabular}} & \multicolumn{1}{l}{\begin{tabular}[c]{@{}r@{}}English\\   Comments\end{tabular}} & \multicolumn{1}{r}{Total} &  &  &  &  \\ \cline{1-4}
\multicolumn{1}{l}{Angry}                                                                   & \multicolumn{1}{r}{167,692}                    & \multicolumn{1}{r}{206,994}                      & \multicolumn{1}{r}{374,686}         &  &  &  &  \\
\multicolumn{1}{l}{Haha}                                                                    & \multicolumn{1}{r}{79,444}                     & \multicolumn{1}{r}{162,149}                      & \multicolumn{1}{r}{214,593}          &  &  &  &  \\
\multicolumn{1}{l}{Wow}                                                                     & \multicolumn{1}{r}{38,433}                     & \multicolumn{1}{r}{61,720}                      & \multicolumn{1}{r}{100,153}          &  &  &  &  \\
\multicolumn{1}{l}{Sad}                                                                     & \multicolumn{1}{r}{28,271}                     & \multicolumn{1}{r}{102,264}                      & \multicolumn{1}{r}{130,535}          &  &  &  &  \\
\multicolumn{1}{l}{Love}                                                                    & \multicolumn{1}{r}{34,019}                     & \multicolumn{1}{r}{300,600}                      & \multicolumn{1}{r}{334,619}          &  &  &  &  \\ \cline{1-4}
\multicolumn{1}{l}{Total}                                                          & \multicolumn{1}{r}{347,859}                    & \multicolumn{1}{r}{833,727}                     & \multicolumn{1}{r}{1,181,586}         &  &  &  &  \\ \cline{1-4}
\end{tabular}
\end{table}

\begin{figure}[h!]
\centering
\includegraphics[scale=0.32]{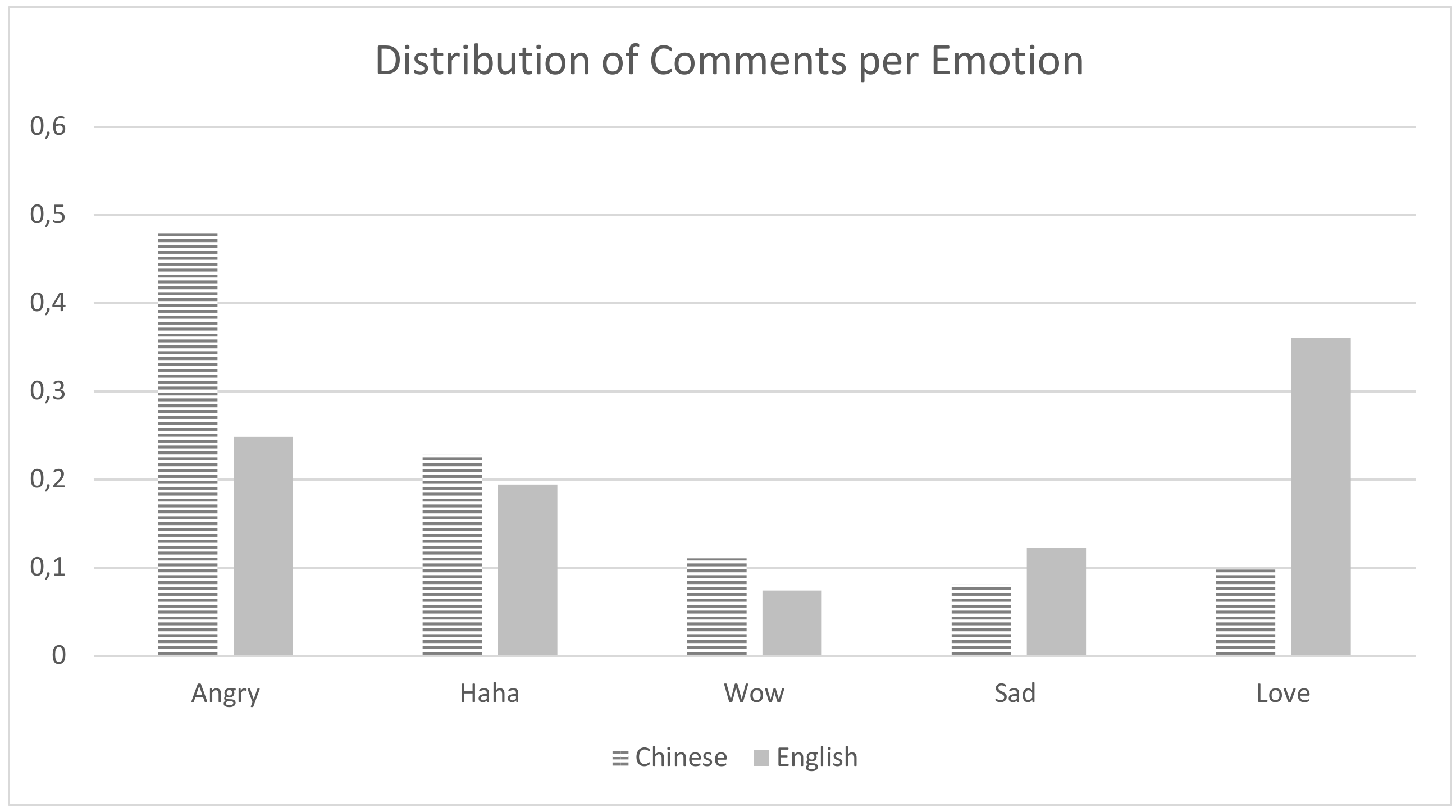}
\caption{Distribution of comments across emotions for English and Chinese.}
\label{fig:distribution}
\end{figure}

The comments in Chinese are in traditional Mandarin characters from predominantly Taiwanese news media.
It is interesting to notice how English posts have a much higher comment density.
The percentage that every reaction represents in the datasets was also calculated.
Results for this distribution are presented in Figure~\ref{fig:distribution}.
It can be observed that the distribution for both Chinese comments and English comments have similar behaviors for the reactions in the middle, but very opposing distributions for Angry and Love.
This kind of distribution can provide some insights on how different groups interact with the platform.
Furthermore, this can lead to a deeper study on the differences or similarities in interaction based on cultural or language backgrounds.Though it must be made clear that this behavior may not be universal and is probably dependant on the time period crawled and the trending news contained in it. 

% \begin{figure}[b!]
% \centering
% \includegraphics[scale=0.32]{annotator_english}
% \caption{Annotator Agreement for English Comments.
% \emph{1 Label} indicates the amount of comments that where labeled as sarcasm by at least one annotator.
% \emph{2 Label} and \emph{3 Label} correspond to cases where at least two and three annotators, respectively, labeled the comment as being sarcastic.}
% \label{fig:annotator_en}
% \end{figure}

% \begin{figure}[t!]
% \centering
% \includegraphics[scale=0.32]{annotator_chinese}
% \caption{Annotator Agreement for Chinese Comments.
% \emph{1 Label} indicates the amount of comments that where labeled as sarcasm by at least one annotator.
% \emph{2 Label} and \emph{3 Label} correspond to cases where at least two and three annotators, respectively, labeled the comment as being sarcastic.}
% \label{fig:annotator_ch}
% \end{figure}

Another factor to take into account is that the data comes from news media posts, therefore there may be a tendency to elicit some reactions more than others. This can explain why anger has a significant share in both languages while Wow and Sad are not so common, perhaps due to the nature of the news shared.

Separate sets of data were crawled at different times to perform evaluation.
For evaluation, human annotation was required.
The following subsection describes the process for ground truth generation.

\begin{figure*}[ht!]
\centering
\includegraphics[scale=0.67]{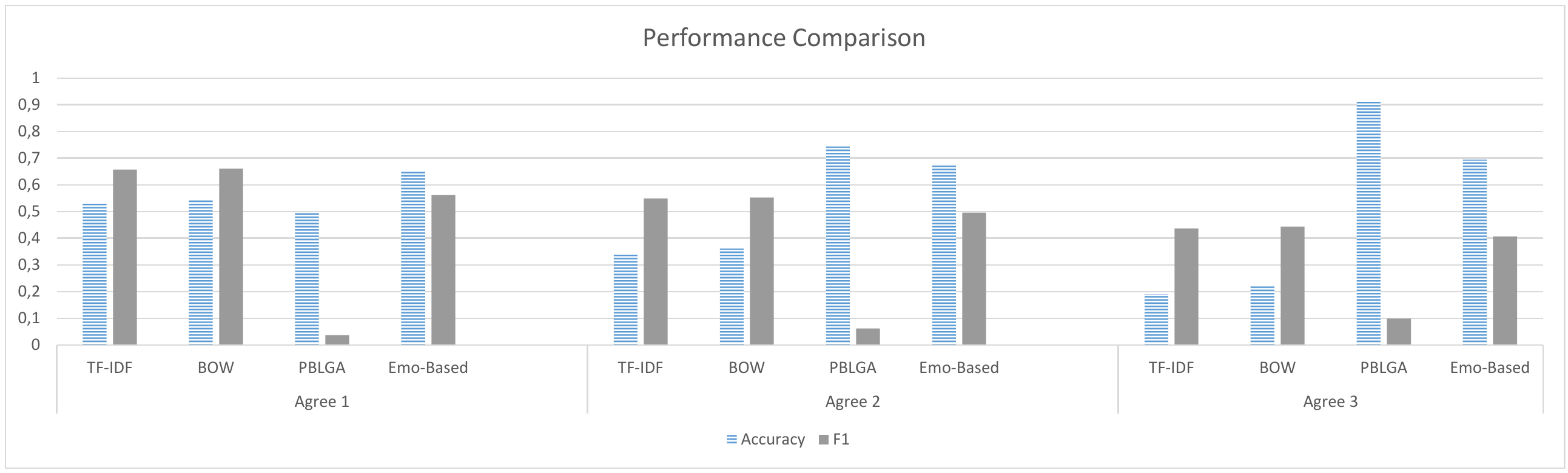}
\caption{Sarcasm classifier performance comparison by Accuracy and F-Measure for English experiments on the Amazon Turk Test set.}
\label{fig:performance}
\end{figure*}

\subsection{Ground Truth}
\label{subsec:groundtruth}

As mentioned previously, one of the main challenges in the sarcasm detection task is the difficulty even for humans to identify sarcastic expressions.
It is required to have annotated testing sets of good quality and consistency in order to perform a proper evaluation. We therefore generated several testing sets, 4 for English evaluation and 3 for Chinese. The inter-annotator consistency of out datasets is measured by Fleiss' Kappa. A good inter-annotator score guarantees the quality of the testing sets, while maintaining this quality across sets is an indicator of consistency. The details of the testing sets are provided in Table~\ref{tab:testing}.

\renewcommand\tabcolsep{1pt}
\begin{table}[b]
\centering
\caption{Details of the testing sets to be used for evaluation.}
\label{tab:testing}
\begin{tabular}{l|c|c|c|c}
\hline
{\textbf{Language}}                                      & \multicolumn{1}{c|}{\textbf{Set}} & \multicolumn{1}{l|}{\textbf{\# of Texts}} & \multicolumn{1}{l|}{\textbf{\# of Annotators}} & \multicolumn{1}{l}{\textbf{Fleiss' Kappa}} \\ \hline
\multicolumn{1}{c|}{{\textbf{English}}} & Test 1                           & 430                                 & 3                                     & 0.7426                            \\
\multicolumn{1}{c|}{}                         & Test 2                           & 260                                 & 4                                     & 0.7391                            \\
\multicolumn{1}{c|}{}                         & Test 3                           & 400                                 & 4                                     & 0.7563                            \\
\multicolumn{1}{c|}{}                         & Test Turk                        & 720                                 & 3                                     & 0.7148                            \\ \hline
{\textbf{Chinese}}                      & Test 1                           & 300                                 & 5                                     & 0.8560                            \\
                                              & Test 2                           & 294                                 & 6                                     & 0.9104                            \\
                                              & Test 3                           & 346                                 & 6                                     & 0.7630                            \\ \hline
\end{tabular}
\end{table}

The annotation task requires the annotators to label a comment as being sarcastic or not. The comments for the English sets are a combination of different platforms and contain Facebook comments as well as Tweets from Twitter. The texts are collected randomly from several time periods to avoid any particular bias to a specific news or season. The comments for the Chinese testing sets are collected only from Facebook comments, but again being posted in different articles and time periods randomly selected.

The annotators for the English and Chinese Test 1, 2 ,3 sets are university students between 22 and 29 years old,  native speakers of the language being evaluated. The subjects are familiar to the social media platforms and the sarcastic posting behavior in it. Not all annotators evaluate all of the sets as observed in the ''\#of Annotators'' column in Table~\ref{tab:testing}, different combinations of annotators worked on different sets but as observed they maintain good Fleiss' Kappa scores. According to the suggested interpretation all sets achieve at least Substantial agreement(0.61-0.80).

\begin{figure}[b!]
\centering
\includegraphics[scale=0.32]{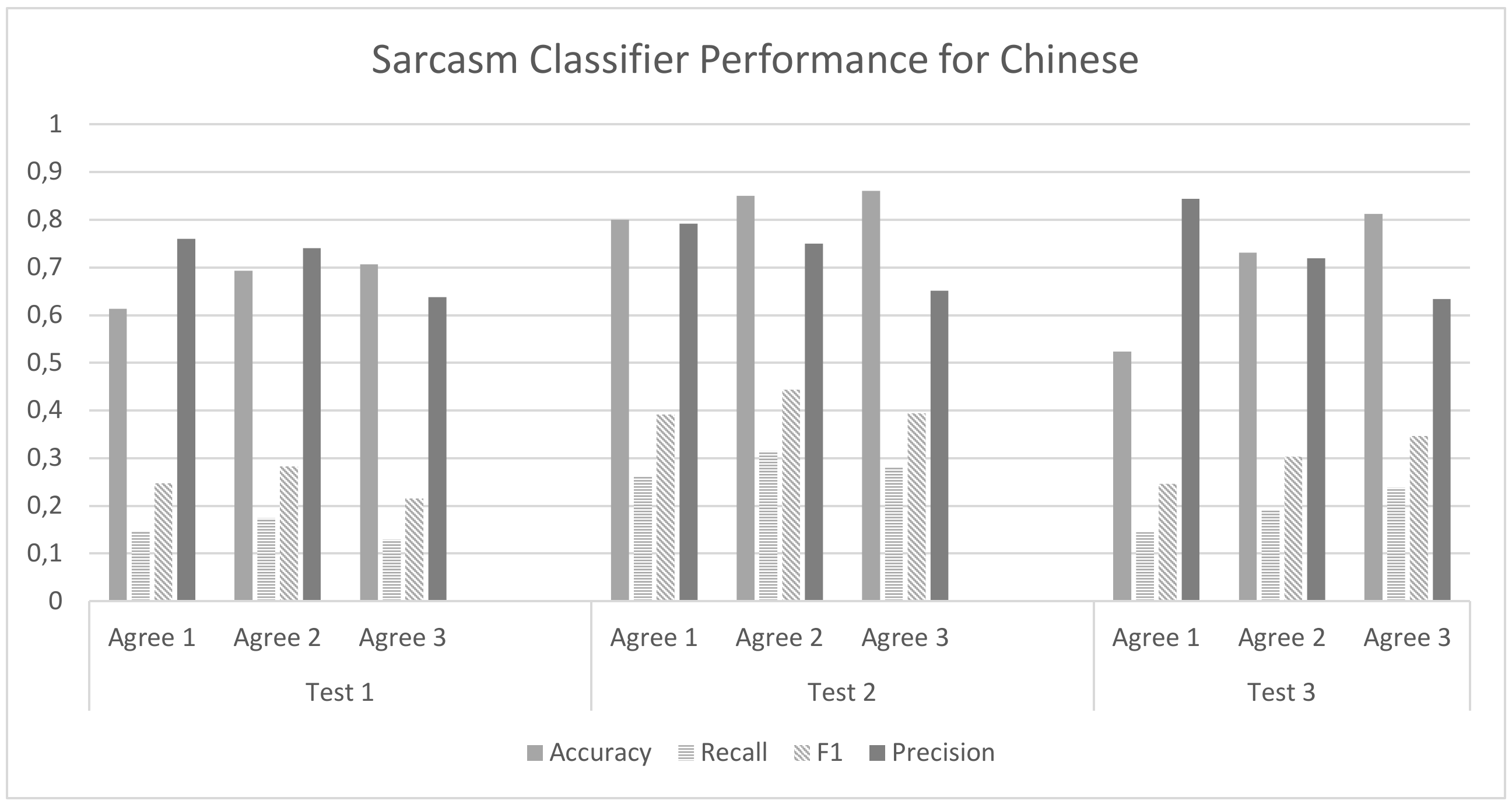}
\caption{Performance of sarcasm classifier for Chinese data illustrated by Accuracy, Recall, F1-Score and Precision metrics.
}
\label{fig:ch_performance}
\end{figure}

% \begin{figure}[b!]
% \centering
% \includegraphics[scale=0.32]{percentage_annotator}
% \caption{Different trends for the percentage of comments from the total set labeled as sarcasm for English and Chinese.
% \emph{1 Label} indicates the amount of comments that where labeled as sarcasm by at least one annotator.
% \emph{2 Label} and \emph{3 Label} correspond to cases where at least two and three annotators, respectively, labeled the comment as being sarcastic.}
% \label{fig:perc_annotator}
% \end{figure}

% Please add the following required packages to your document preamble:
% \usepackage{multirow}

The Test Turk set comes from a task submitted to Amazon Mechanical Turk where every text was rated by 3 annotators. Additionally they were asked to provide a degree of intensity which is not used in this work but might come useful in the future. The task contained a few manually inserted comments regarded as definitely sarcastic and definitely not sarcastic to verify if the annotators could perform the evaluation correctly.

\subsection{Evaluation}
\label{subsec:evaluation}

\subsubsection{English Data Test}
To evaluate the performance of our Sarcasm Classifier for English texts three comparison methods were implemented. The first two are text classification baselines trained with a corpus related to the topic at hand using TF-IDF and Bag of Words(BOW) based features respectively to train Na\"ive Bayes classifiers. Both classifiers were trained using 2400 short documents, 1200 being sarcastic and the other 1200 with no presence of sarcasm. The third comparison method is an implementation of the Parsing-Based Lexicon Generation Algorithm(PBLGA) method developed by Bharti et al.~\cite{bharti2015parsing}. This method was trained with 40,000 short documents containing the hashtag \#sarcasm as indicated by the referenced work. The method introduced in this work will be referred to as Emo-Based in the results to be presented.

All four English test sets were processed by the four classifiers mentioned in the previous paragraph. Tree different levels of agreement are considered to determine the correctness of a classification. \textbf{Agree 1} means that the output label of the classifier matched the label of at least 1 annotator. \textbf{Agree 2} requires the classifier to match the label of at least 2 human annotators. Subsequently \textbf{Agree 3} means at least 3 annotators agree to a label and so does the classifier. Intuitively Agree 1 will contain more texts regarded as sarcasm by the annotation since it only requires one annotator to label it as sarcasm, Agree 3 on the other hand will contain less cases of sarcasm since 3 annotators need to agree on it which makes it a more strict policy.
% The sets of comments that were used for annotation were processed in parallel by the automatic detection system.
% After performing the evaluation, the system simply assigns a label of sarcasm or no sarcasm to every processed comment.
% For each of the ground truths, a result is generated whenever there is a match between the system and the annotation label.
% Agree 1 corresponds to a match between the system's result and \emph{1 Label} annotation, Agree 2 and 3 correspond to \emph{2} and \emph{3 Label}, respectively.
% The performance metrics considered are Precision, Recall and F-Measure.

Figure~\ref{fig:performance} presents accuracy and F1-score performance for all classifiers across the three described levels of agreement for the Amazon Turk test set. It can be observed that the Emo-Based method proposed in this work has a more stable performance across levels of agreement, specially regarding accuracy. TF-IDF and BOW methods perform well in the Agree 1 evaluation since they are content based methods and the first level of agreement contains more texts labeled as sarcasm. But their performance is affected as the ground truth gets more strict. After taking a look at the classification by these methods we found they are very generous in assigning a sarcasm label since it is determined by the presence of specific terms but doesn't consider context. 

PBLGA method on the other hand doesn't perform well in terms of F1 score but improves in accuracy with the level of agreement which appeared surprising. After taking a look at the classification output of PBLGA it was found that opposite from the content based methods, PBLGA is very selective on where to label sarcasm, therefore the fewer cases of sarcasm present in the ground truth, the better the accuracy since it will classify most of the non-sarcasm correctly, nevertheless it suffers alot in recall. The Emo-Based sarcasm classifier receives emotions from a pattern based approach which can provide more context. Additionally the candidate filtering process as well as the Distance Ratio and Score Ratio measurements introduced in the methodology make the classifier not so generous yet at the same time not overly selective when labeling a text as sarcasm. A complete detail of performance results across all data sets and including additional indicators as Recall and Precision is presented in Table~\ref{tab:results}.

The results reflect a more consistent performance from the Emo-Based method across datasets and agreement levels. It can be observed that it dominates the Accuracy for Agree 1 on all data sets, more importantly this performance is maintained despite the ground truth becoming more strict. The proposed method also dominates precision across the board reflecting a good performance of the filtering process defined in the methodology. Other methods tend to lean to one class, which in the case of TF-IDF and BOW favors their recall. It can also be observed that PBLGA method presents no score for F1 and recall in some instances. This was analyzed into more detail and it was found that PBLGA labeled very few texts as sarcasm and as the agree level increases also fewer sarcasm instances remain, the 0 scores in the table result of there being no matching between these scarce labels from the classifier and the ground truth.

\begin{table*}[t!]
\centering
\caption{Performance comparison against other methods for different data sets and varying levels of annotator agreement.}
\label{tab:results}
\begin{tabular}{ll|rrrr|rrrr|rrrr}
\hline
                                                      &                             & \multicolumn{4}{c|}{\textbf{Agree 1}}                                                                                                                   & \multicolumn{4}{c|}{\textbf{Agree 2}}                                                                                                                   & \multicolumn{4}{c}{\textbf{Agree 3}}                                                                                                                   \\ \hline
\multicolumn{1}{c|}{Set}                              & \multicolumn{1}{c|}{Method} & \multicolumn{1}{c}{\textit{Accuracy}} & \multicolumn{1}{c}{\textit{F1}} & \multicolumn{1}{c}{\textit{Recall}} & \multicolumn{1}{c|}{\textit{Precision}} & \multicolumn{1}{c}{\textit{Accuracy}} & \multicolumn{1}{c}{\textit{F1}} & \multicolumn{1}{c}{\textit{Recall}} & \multicolumn{1}{c|}{\textit{Precision}} & \multicolumn{1}{c}{\textit{Accuracy}} & \multicolumn{1}{c}{\textit{F1}} & \multicolumn{1}{c}{\textit{Recall}} & \multicolumn{1}{c}{\textit{Precision}} \\ \hline
\multicolumn{1}{l|}{{\textbf{Test 1}}} & \textit{\textbf{Emo-Based}} & \textit{\textbf{0,6023}}              & 0,4896                          & 0,3814                              & \textbf{0,6833}                         & \textit{0,6535}                       & 0,4494                          & 0,3840                              & \textbf{0,5417}                         & \textit{0,7047}                       & 0,4227                          & 0,4286                              & \textbf{0,4170}                        \\
\multicolumn{1}{l|}{}                                 & \textit{\textbf{PBLGA}}     & 0,4907                                & 0,0179                          & 0,0093                              & 0,2500                                  & 0,7000                                & 0,0301                          & 0,0160                              & 0,2500                                  & 0,8674                                & 0,0000                          & 0,0000                              & 0,1650                                 \\
\multicolumn{1}{l|}{}                                 & \textit{\textbf{TFIDF}}     & 0,5140                                & 0,6579                          & 0,9349                              & 0,5076                                  & 0,3186                                & 0,5539                          & 0,9120                              & 0,3977                                  & 0,1651                                & 0,4465                          & 0,8776                              & 0,2994                                 \\
\multicolumn{1}{l|}{}                                 & \textit{\textbf{BOW}}       & 0,5209                                & 0,6544                          & 0,9070                              & 0,5118                                  & 0,3349                                & 0,5502                          & 0,8800                              & 0,4003                                  & 0,1953                                & 0,4463                          & 0,8571                              & 0,3017                                 \\ \hline
\multicolumn{1}{l|}{{\textbf{Test 2}}} & \textit{\textbf{Emo-Based}} & \textit{\textbf{0,6000}}              & 0,5000                          & 0,3939                              & \textbf{0,6842}                         & \textit{0,6769}                       & 0,4891                          & 0,4429                              & \textbf{0,5461}                                  & \textit{0,6615}                       & 0,3528                          & 0,3125                              & \textbf{0,4051}                        \\
\multicolumn{1}{l|}{}                                 & \textit{\textbf{PBLGA}}     & 0,4923                                & 0,0149                          & 0,0076                              & 0,5000                                  & 0,7231                                & 0,0000                          & 0,0000                              & 0,2500                                  & 0,8692                                & 0,0000                          & 0,0000                              & 0,1650                                 \\
\multicolumn{1}{l|}{}                                 & \textit{\textbf{TFIDF}}     & 0,5385                                & 0,6512                          & 0,8485                              & 0,5283                                  & 0,3692                                & 0,5456                          & 0,8429                              & 0,4033                                  & 0,2846                                & 0,4650                          & 0,9063                              & 0,3127                                 \\
\multicolumn{1}{l|}{}                                 & \textit{\textbf{BOW}}       & 0,5423                                & 0,6469                          & 0,8258                              & 0,5317                                  & 0,3808                                & 0,5408                          & 0,8143                              & 0,4049                                  & 0,3115                                & 0,4679                          & 0,9063                              & 0,3153                                 \\ \hline
\multicolumn{1}{l|}{{\textbf{Test 3}}} & \textit{\textbf{Emo-Based}} & \textit{\textbf{0,6075}}              & 0,5341                          & 0,4545                              & 0,6475                                  & \textit{0,6150}                       & 0,4710                          & 0,4370                              & \textbf{0,5108}                                  & \textit{0,6050}                       & 0,3726                          & 0,3538                              & 0,3934                                 \\
\multicolumn{1}{l|}{}                                 & \textit{\textbf{PBLGA}}     & 0,5125                                & 0,0580                          & 0,0303                              & 0,6667                                  & 0,6950                                & 0,0480                          & 0,0252                              & 0,5000                                  & 0,8300                                & 0,0836                          & 0,0462                              & 0,4433                                 \\
\multicolumn{1}{l|}{}                                 & \textit{\textbf{TFIDF}}     & 0,5075                                & 0,6359                          & 0,8687                              & 0,5015                                  & 0,3500                                & 0,5419                          & 0,8487                              & 0,3980                                  & 0,2700                                & 0,4712                          & 0,8923                              & 0,3201                                 \\
\multicolumn{1}{l|}{}                                 & \textit{\textbf{BOW}}       & 0,5125                                & 0,6314                          & 0,8434                              & 0,5045                                  & 0,3650                                & 0,5387                          & 0,8235                              & 0,4003                                  & 0,3000                                & 0,4751                          & 0,8923                              & 0,3238                                 \\ \hline
\multicolumn{1}{l|}{{\textbf{Turk}}}   & \textit{\textbf{Emo-Based}} & \textit{\textbf{0,6528}}                       & 0,5614                          & 0,4444                              & \textbf{0,7619}                         & \textit{0,6778}                       & 0,4956                          & 0,4389                              & \textbf{0,5690}                                  & \textit{0,6944}                       & 0,4067                          & 0,4038                              & \textbf{0,4096}                        \\
\multicolumn{1}{l|}{}                                 & \textit{\textbf{PBLGA}}     & 0,4972                                & 0,0372                          & 0,0194                              & 0,4375                                  & 0,7444                                & 0,0616                          & 0,0333                              & 0,4063                                  & 0,9139                                & 0,0983                          & 0,0577                              & 0,3319                                 \\
\multicolumn{1}{l|}{}                                 & \textit{\textbf{TFIDF}}     & 0,5292                                & 0,6572                          & 0,9028                              & 0,5167                                  & 0,3375                                & 0,5485                          & 0,9222                              & 0,3903                                  & 0,1903                                & 0,4366                          & 0,9423                              & 0,2841                                 \\
\multicolumn{1}{l|}{}                                 & \textit{\textbf{BOW}}       & 0,5444                                & 0,6605                          & 0,8861                              & 0,5264                                  & 0,3611                                & 0,5527                          & 0,9056                              & 0,3977                                  & 0,2222                                & 0,4435                          & 0,9423                              & 0,2900                                 \\ \hline
\end{tabular}
\end{table*}

\subsubsection{Chinese Data Test}
To test the multilingual capabilities of our method, a Chinese classifier was implemented as defined in our methodology. The classifier performance was evaluated across the three testing sets mentioned in Subsection~\ref{subsec:groundtruth}. Figure~\ref{fig:ch_performance} presents the results for classification of sarcasm in Chinese comments on all standard metrics. Similar as in English, the classifier achieves better scores for accuracy and precision. This behavior is sustained across different levels of agreement and different testing sets which again is an indicator of a well balanced classifier.
The importance of these results is that the presented method learns directly from the data, no external human knowledge is added. Although the performance metrics don't present very high scores it still provides a sense that the features evaluated in this work can indeed provide some clues for sarcastic posts.

\section{Conclusions and Future Work}

This work tries to bring focus to the importance of understanding the platforms being used, how users interact in them, and, more importantly, how we can make use of these behaviors when working towards identifying sarcasm.
In this particular case, background knowledge of Facebook pages from news media gave some particular insights that later played an important role in the development of the method.
Some examples include common behaviors of internet trolls, the usage of ``Reactions'' buttons, and other commenting tendencies.
There is still much to be done in terms of developing precise, efficient, and effective methods for sarcasm detection.
To the best of our knowledge, this is the first work using Facebook reactions as emotion signals for any sentiment-related task. We do not intend to set state of the art performance, but to provide some evidence that the features evaluated in this work can be useful in the task of sarcasm classification. More importantly defining a method where these feature can be learned by the system without external human knowledge.

This work also provides a brief view on some behaviors related to the data and cultural and language differences.
The extracted emotion patterns can serve as a summary of the data being dealt with and can also intuitively reflect many characteristics of the audience based on the language or expressions they use.
It is also interesting to notice that the differences of emotion reactions in the data collection.
Different languages reflect different uses of emotions in posts, as it was observed with the higher percentage of love posts by English users than those in Chinese.
Although it is still risky to call it a cultural difference, it can be assumed that there was a situational difference where at the time one of the language groups was more leaned towards those emotions given the ongoing events.

Extensions of this work will attempt to improve the performance metrics where it stayed behind.
In order to obtain a higher recall, for instance, more examples of sarcastic comments(explicit and not explicit) may be used in the training process for the patterns to learn their characteristics.
More context can be included in the analysis, for example the elicited polarity of the original news post.
Additionally, other data sources must be considered, since some of the behaviors are very particular to news sites; this might pose a difficulty when trying to perform an extensive study.
Data is already being collected in Spanish to develop a similar system and test if the method is indeed multi-lingual, or to evaluate to what degree it is.
Additional studies that can be derived from this work include the study of cultural differences in sarcastic posting behavior.
Likewise, evaluating which kind of posts are more prone to receive sarcastic replies can also be carried out.
Finally, it is a possibility to study the role of language in the usage, understanding, and proliferation of sarcasm.

\bibliographystyle{ACM-Reference-Format}
\bibliography{webconf2018bib} 

\end{document}